\documentclass[pdflatex,sn-mathphys-num]{sn-jnl}% Math and Physical Sciences Numbered Reference Style
\usepackage{graphicx}%
\usepackage{multirow}%
\usepackage{amsmath,amssymb,amsfonts}%
\usepackage{amsthm}%
\usepackage{mathrsfs}%
\usepackage[title]{appendix}%
\usepackage{xcolor}%
\usepackage{textcomp}%
\usepackage{manyfoot}%
\usepackage{booktabs}%
\usepackage{algorithm}%
\usepackage{algorithmicx}%
\usepackage{algpseudocode}%
\usepackage{listings}%
\usepackage{longtable}
\usepackage{booktabs}
\usepackage{caption}
\usepackage{hyperref}
\usepackage{setspace}
\usepackage{pifont}
\usepackage{times} % Times New Roman font
\usepackage{framed}
\usepackage{tabularx}
\usepackage{mdframed}
\usepackage{array}

\theoremstyle{thmstyleone}%
\theoremstyle{thmstyletwo}%

\theoremstyle{thmstylethree}%

\begin{document}

% \title[AT]{BioMol-LLM-Bench: Systematic cross-scale evaluation of large language models for bio-molecular research}
\title[AT]{The limits of bio-molecular modeling with large language models : a cross-scale evaluation}    % in molecular understanding}
% The limits of bio-molecular modeling with large language models : a cross-scale evaluation

%%=============================================================%%
%% GivenName	-> \fnm{Joergen W.}
%% Particle	-> \spfx{van der} -> surname prefix
%% FamilyName	-> \sur{Ploeg}
%% Suffix	-> \sfx{IV}
%% \author*[1,2]{\fnm{Joergen W.} \spfx{van der} \sur{Ploeg} 
%%  \sfx{IV}}\email{iauthor@gmail.com}
%%=============================================================%%

\author[1,2]{\fnm{Yaxin Xu}}\email{xuyx2024@mail.sustech.edu.cn}
\equalcont{These authors contributed equally to this work.}

\author[2]{\fnm{Yue Zhou}}\email{zhouy@pcl.ac.cn}
\equalcont{These authors contributed equally to this work.}

\author[3]{Tianyu Zhao\,\orcid{https://orcid.org/0000-0003-3121-6042}} \email{tyzhao@mail.bnu.edu.cn}
% \author[3]{Tianyu Zhao} \email{tyzhao@mail.bnu.edu.cn}

\author[1]{Fengwei An} \email{anfw@sustech.edu.cn}

\author*[2]{Zhixiang Ren\,\orcid{https://orcid.org/0000-0002-4104-3790}}\email{jason.zhixiang.ren@outlook.com}

% \affil[1]{\orgdiv{Department}, \orgname{Organization}, \orgaddress{\street{Street}, \city{City}, \postcode{10587}, \state{State}, \country{Country}}}

\affil[1]{\orgname{Southern University of Science and Technology}, 
          \orgaddress{\city{Shenzhen}, \postcode{518055}, \country{China}}}
          
\affil*[2]{\orgdiv{Pengcheng Laboratory}, \orgaddress{\city{Shenzhen}, \postcode{518055}, \country{China}}}

\affil[3]{Institute of Mechanics, Chinese Academy of Sciences, Beijing, 100190, China}

% \affil[3]{\orgdiv{Department}, \orgname{Organization}, \orgaddress{\street{Street}, \city{City}, \postcode{610101}, \state{State}, \country{Country}}}

%%==================================%%
%% Sample for unstructured abstract %%
%%==================================%%

\abstract{
% Large language models are increasingly applied to bio-molecular discovery, yet systematic evaluation across the multi-scale hierarchy of biological problems—from small molecules to protein complexes—remains limited.
% Large language models are increasingly being applied to bio-molecular discovery, yet systematic evaluation across multi-scale biological problem levels, as well as assessment of models' tool-augmented capabilities remain limited.
The modeling of bio-molecular system across molecular scales remains a central challenge in scientific research.
Large language models (LLMs) are increasingly applied to bio-molecular discovery, yet systematic evaluation across multi-scale biological problems and rigorous assessment of their tool-augmented capabilities remain limited.
% 工具的痛点
% Here we present the cross-scale bio-molecular benchmark, a unified framework comprising 26 downstream tasks spanning prediction and generation tasks. across four levels of input complexity, with integrated tool interfaces enabling agentic workflows.
We reveal a systematic gap between LLM performance and mechanistic understanding through the proposed cross-scale bio-molecular benchmark: BioMol-LLM-Bench, a unified framework comprising 26 downstream tasks that covers 4 distinct difficulty levels, and computational tools are integrated for a more comprehensive evaluation.
% 强调数据全 数据集亮点
Evaluation on 13 representative models reveals 4 main findings: chain-of-thought data provides limited benefit and may even reduce performance on biological tasks; hybrid mamba–attention architectures are more effective for long bio-molecular sequences; supervised fine-tuning improves specialization at the cost of generalization; and current LLMs perform well on classification tasks but remain weak on challenging regression tasks. Together, these findings provide practical guidance for future LLM-based modeling of molecular systems.

}


%%\pacs[JEL Classification]{D8, H51}

%%\pacs[MSC Classification]{35A01, 65L10, 65L12, 65L20, 65L70}

\maketitle

% Main content
\section{Introduction}
\label{sec:introduction}

Bio-molecule plays a foundational role across a wide range of biological and chemical domains\cite{kortemme2004ppibaker, gibson2008syn, powner2009synthesis}. 
Multi-scale bio-molecular modeling has emerged as a critical paradigm, aiming to bridge different levels of representation from monomer-level molecule to polymer complex. 
% This shift enables the joint consideration of structural, thermodynamic, and functional properties, providing a more holistic understanding of polymer systems. 
They underpin applications from bio-materials design to drug delivery systems\cite{douglas2012logic, langer2004designing}. 
However, effectively integrating and modeling information across these scales remains a significant challenge.
% With the rapid growth of data availability and computational power, there is an increasing reliance on data-driven and deep learning models to accelerate property prediction and model complex interactions. 
With the rapid growth of data availability and computational power, there is an increasing reliance on large language models (LLMs)\cite{achiam2023gpt, team2023gemini, team2024gemma, liu2024deepseek, bai2023qwen, touvron2023llama, basant2025nemotron, abdin2024phi4} to accelerate property prediction and model complex interactions simultaneously. 
% Real-world polymer systems are inherently complex, often involving multiple categories of polymers such as homopolymers, copolymers, and blends, as well as diverse interaction types including polymer–polymer, polymer–solvent, and polymer–additive interactions. 
% These complexities are central to industrial applications, where optimizing performance metrics such as mechanical strength, thermal stability, and solubility must be balanced with considerations of processability and manufacturability. 
% From a scientific perspective, this introduces substantial challenges due to strong cross-scale dependencies, where molecular-level features influence mesoscopic organization and ultimately macroscopic properties. 
% Consequently, accurate modeling requires moving beyond isolated property prediction toward coupled, multi-objective learning frameworks that can capture interdependent behaviors across scales.
% The rapid evolution of large language models (LLMs) has transformed artificial intelligence, enabling unprecedented capabilities in natural language understanding, generation, and reasoning\cite{achiam2023gpt, team2023gemini, team2024gemma, liu2024deepseek, bai2023qwen, touvron2023llama, anil2023palm, glm2024chatglm, basant2025nemotron, abdin2024phi4}.
% Particularly, initially developed for general text processing, large language models (LLMs)\cite{achiam2023gpt, team2023gemini, team2024gemma, liu2024deepseek, bai2023qwen, touvron2023llama, anil2023palm, glm2024chatglm, basant2025nemotron, abdin2024phi4} have increasingly been adapted to specialized scientific domains. 
In bio-molecular sciences, these models\cite{bran2023chemcrow, beltagy2019scibert} hold particular promise: the ability to reason about molecules, proteins, and their interactions in natural language, enable more intuitive exploration of chemical space.
%, and democratize access to specialized computational methods. 
Recent years have witnessed a proliferation of domain-adapted models\cite{wang2025txgemma, pei2024biot5+, xia2025naturelm, zhuang2025instructbiomol}, ranging from general-purpose foundation models fine-tuned on scientific corpora to architectures specifically designed for bio-molecular understanding, each claiming varying degrees of proficiency on bio-molecular tasks.
Notable examples include TxGemma\cite{wang2025txgemma}, which specializes in therapeutic tasks by processing diverse modalities such as small molecules and natural text to predict therapeutic properties. 
% Similarly, NatureLM\cite{xia2025naturelm} extends this paradigm as a sequence-based foundation model by unifying multiple types of bio-molecules to support protein design, drug optimization, and cross-domain design tasks. 

% Limitations of Existing Benchmarks
Yet the rapid development of these models has outpaced our ability to systematically evaluate their capabilities. 
General-domain benchmarks\cite{rein2024gpqa, wang2024mmlu, wang2023scibench, sun2024scieval, olea2024ai2arc, phan2025hle, saikh2022scienceqa, laurent2024labbench} such as MMLU-Pro\cite{wang2024mmlu} and AI2ARC\cite{olea2024ai2arc}, while useful for assessing broad reasoning capabilities, lack the specialized knowledge required for meaningful evaluation on bio-molecular tasks. 
% Existing benchmarks, while valuable, suffer from critical limitations that constrain their utility for guiding model development and deployment decisions.
Conversely, domain-specific benchmarks\cite{shen2024proteinlmbench, walker2010chembench, yu2024llasmol, wu2018moleculenet, zhu2023marcel, rao2019tape, dallago2021flip} often focus on single task types such as molecular reaction outcome prediction, or protein function annotation, without considering how models perform across the interconnected scales of bio-molecular problems. 
These limitations obscure important patterns in model behavior: a model that excels at predicting small molecule solubility may fail catastrophically on protein-protein interaction tasks, such trade-offs remain poorly characterized.
More significantly, current evaluation frameworks\cite{gao2025txagent, ding2025scitoolagent} provide limited insight into how different model architectures and training approaches affect model behavior across cross-scale bio-molecular tasks. 
% Many focus narrowly on individual task categories, failing to capture the multi-scale nature of bio-molecular problems that span small molecules, peptides, proteins, and their complex interactions. 
% Others evaluate models in isolation, neglecting the growing reality that production systems increasingly augment language models with external computational tools. 

% 描述上 明确说明痛点
Furthermore, LLM is evolving towards agents and absorbing tool retrieval capabilities. 
Existing benchmarks\cite{notin2023proteingym, thumuluri2022deeploc2, huang2020deeppurpose, zhao2025abbibench, sorkun2019aqsoldb, li2024prostage} lack integration with computational tools and are limited to task-oriented question answering.
But in real-world scientific practice, researchers combine conceptual understanding with specialized software\cite{gao2025tooluniverse, ding2025scitoolagent, bran2023chemcrow, abramson2024af3, swanson2024admet, van2024foldseek} for bio-molecular modeling. 
Benchmarks that mentioned above typically evaluate models under artificial constraints, requiring direct question answering or chain-of-thought (CoT) thinking, thereby failing to provide a comprehensive assessment of model capabilities.
% therefore provides an incomplete picture of their potential utility as components of agentic workflows.

\begin{figure}[h]   %[htbp]
    \centering
    \includegraphics[width=1.0\textwidth]{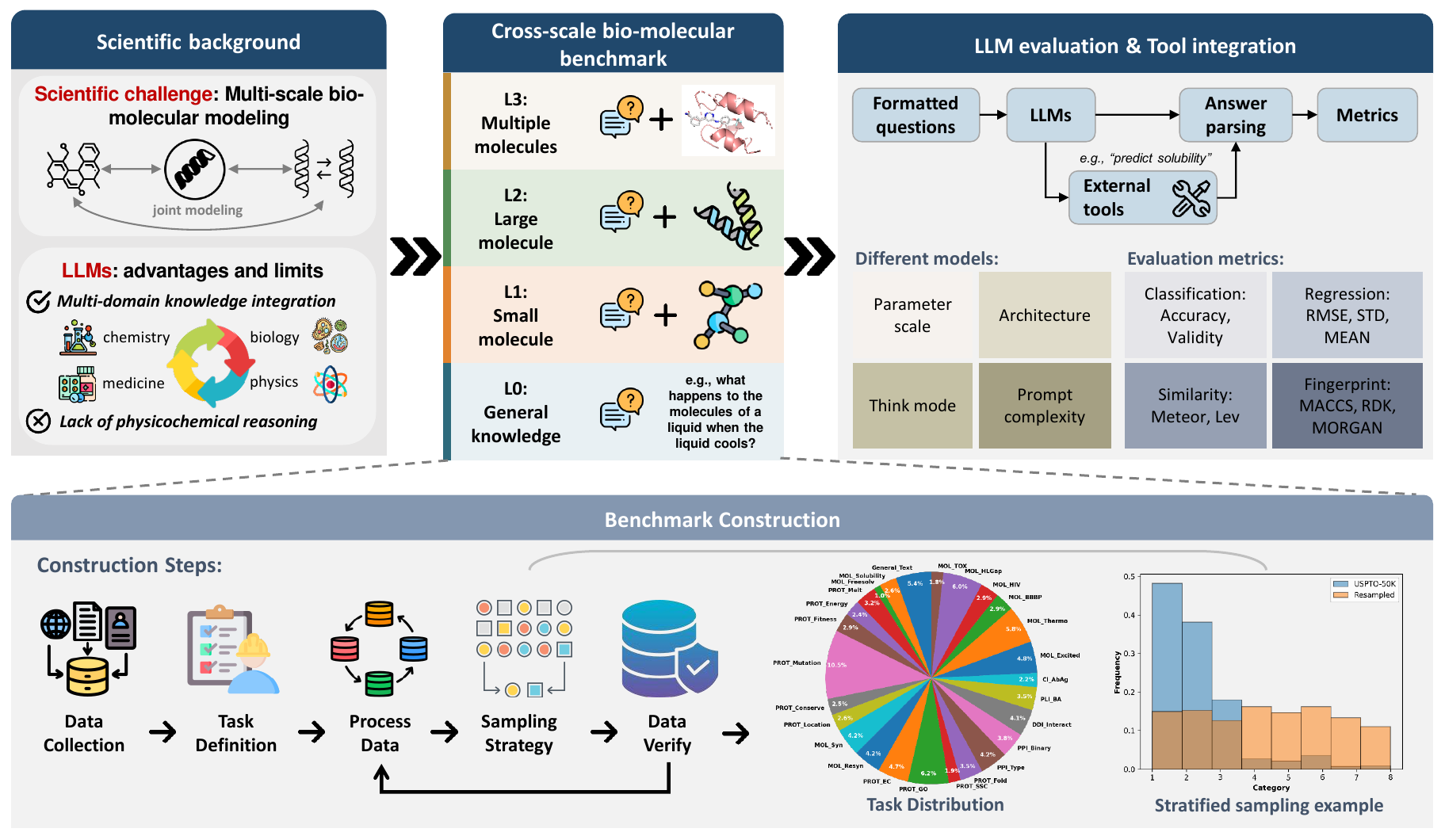}
    \caption{\textbf{Overview of BioMol-LLM-Bench.}
        The framework addresses limitations in existing benchmarks by incorporating specialized bio-chemistry tools and enabling cross-scale evaluation with 26 tasks across 4 levels. 
        The evaluation pipeline integrates automatic answer extraction and metrics computation.
        13 models are compared under different experiment settings.}
    \label{fig:fig1}
\end{figure}

% The Cross-Scale BioMolecular Benchmark: Design Principles and Construction
To address these gaps, we introduce the cross-scale bio-molecular benchmark (BioMol-LLM-Bench), a comprehensive evaluation framework (Figure \ref{fig:fig1}) that aims to help us evaluate the capabilities of LLMs from the perspective of practical scientific applications. Our benchmark encompasses 4 hierarchical levels ($L_0$ to $L_3$) corresponding to increasing structural and functional complexity. This hierarchical organization enables fine-grained analysis of how model capabilities scale with problem complexity. We provide an automated evaluation pipeline that parses model outputs into required formats. Beyond traditional accuracy measures, the pipeline evaluates output validity, which determines whether model responses conform to required formats. Specialized computational tools are integrated into the evaluation framework. This allows evaluation of model ability in computational tool orchestrating and argument parsing.
% designed around three core principles: multi-scale coverage, automatic evaluation, and tools integration.
% The construction of BioMol-LLM-Bench followed a rigorous multi-stage curation and evaluation pipeline designed to ensure data quality and bio-molecular relevance. 
% \begin{itemize}
%     \item Our benchmark encompasses four hierarchical levels ($L_0$ to $L_3$) corresponding to increasing structural and functional complexity.
%     This hierarchical organization enables fine-grained analysis of how model capabilities scale with problem complexity.
%     \item We provide an automated evaluation pipeline that parses model outputs into required formats.
%     % For different task types, we design and apply distinct evaluation metrics.
%     Beyond traditional accuracy measures, the pipeline evaluates output validity, which determines whether model responses conform to required formats.
%     \item Specialized computational tools are integrated into the evaluation framework.
%     This allows evaluation of model ability in computational tool orchestrating and argument parsing.
% \end{itemize}

Furthermore, we conducted experiments on 13 general-purpose and domain-specific models, revealing the benchmark's discriminative power and utility in guiding future LLM design.
\begin{itemize}
    \item \textbf{Training data:} On biological tasks, we found that CoT provides a slight improvement and may even weaken performance. The reasoning process appears sound in linguistics, but contains self-contradictory description and chemical inconsistency.
    \item \textbf{Model Architecture :} We identify that hybrid mamba-attention architectures may offer advantages over pure transformer in bio-molecular sequences processing with long-range dependencies, which even outperforms models with 10 times more parameters.
    % \item On biological tasks, we found that CoT provides a slight improvement to LLMs and may even weaken its performance.  
    %, especially in generation settings, where it often produces biochemically inconsistent reasoning steps and lower-quality outputs.
    % \item We reveal that careful model selection are needed, as performance varies across tasks and exhibits sensitivity to prompt complexity, reasoning mode, and chain-of-thought fine-tuning.
    \item \textbf{Training strategy:} Supervised fine-tuned (SFT) models tend to become narrow specialists and lack generalization ability, which perform poorly on out-of-distribution tasks.
    % We demonstrate that parameter scale and training data diversity remain critical for achieving broad-spectrum scientific reasoning capabilities.
    \item \textbf{Task performance:} Although LLMs perform well on several classification tasks, none of the evaluated models achieves meaningful performance on challenging regression tasks such as amino acid-level property prediction.
    
\end{itemize}

\section{Results}
\label{sec:results}

\subsection{Cross-Scale Bio-molecular Benchmark}

\subsubsection{Benchmark Construction and Data Curation}

The construction of BioMol-LLM-Bench followed a multi-stage pipeline designed to aggregate and refine high-quality bio-molecular data from heterogeneous sources including MMLU-Pro\cite{wang2024mmlu}, USPTO-50K\cite{uspto}, DrugBank\cite{knox2024drugbank}, PDB\cite{pdb}, et. al. 
During the initial data processing phase, raw data were subjected to rigorous deduplication and structural validation to ensure the integrity of SMILES strings and FASTA sequences.
This was further refined through a LLM verification process using a specialized prompt to assign confidence scores ($0 \leq c \leq 1$) to eliminate entries that unrelated to the bio-molecular field. 
% Then, sequence length limits were enforced, and variable dimensions were unified to maintain computational consistency across disparate data formats. 
Then, a hierarchical downstream task stratification was implemented, categorizing tasks into 4 levels ($L_0$ to $L_3$) according to input data type, ranging from general molecular text to specialized large-molecule complexes (Table \ref{tab:tab1}).
Finally, a stratified sampling strategy was applied to the curated dataset of each task to mitigate class imbalance and ensure a representative distribution across the bio-chemical space.
% introduce more in detail.
% add additional meta data, pdb file provided.
% detailed description, total task number, input type, table 1, data process procedure standard given in code

\subsubsection{Integration of Computational Tools}
Current artificial intelligence models, particularly LLMs, possess varying degrees of capability in automatically invoking tools and parsing function parameters.
To further enhance the evaluation capability of the benchmark for models, we integrated a suite of domain-specific tool interfaces\cite{gao2025tooluniverse}. 
% including the PubChem Tool for chemical informatics, the Clinical Tool for trial metadata, the Drug Tool for pharmacokinetic properties, and the UniProt Tool for proteomic data. 
This integration allows for an agentic workflow where models can invoke specific predictive functions, such as $solubility\_lipophilicity\_hydration()$ and $BBB\_penetrance()$, to derive ADMETAI (Absorption, Distribution, Metabolism, Excretion, Toxicity, and AI-predicted) properties.
% By extending with these external computational resources, the benchmark could evaluate the model's ability to interface with real-world scientific tools to solve complex, multi-step biological problems.

\subsubsection{Evaluation Framework and Metrics}

The final stage of the benchmark involves a standardized evaluation pipeline where formatted questions are processed by models under various experimental configurations.
An automatic answer extraction method was proposed to parse model output, so that the extracted content could be used to compare with corresponding labels.
For different downstream tasks, model performance is quantified using distinct metric clusters. 
% Traditional classification and regression tasks are evaluated with accuracy, validity, root mean square error (RMSE), standard deviation (STD), and mean error (MEAN). 
% Generative tasks are assessed via Meteor and Levenshtein distance, while chemical fidelity is strictly monitored through fingerprinting algorithms such as MACCS, MORGAN, and RDK to ensure structural alignment between model outputs and ground-truth molecular topologies.

% introduce in detail.
% released database and code base

\begin{table}[h]
    \centering
    \caption{\textbf{Summary of benchmark tasks.} 
    The table presents 26 tasks categorized into 4 hierarchical levels (L0-L3) based on input complexity.
    For each task, the table specifies the task abbreviation, input modalities, number of test samples (\#Num), and the original data sources.
    SMILES and AA-Seq. represent sequence strings of molecule and protein, respectively.}
    % \begin{tabular}{lcccc}
    \begin{tabularx}{\textwidth}{p{1.2cm}p{3.2cm}p{4.5cm}p{1.2cm}p{4cm}}
        \toprule
        Level & Task (abbr.) & Inputs & \#Num & Source\\
        \midrule
        L0 & General\_Text  & Text & 226 & GPQA\_Diamond\cite{rein2024gpqa}, MMLU\_Pro\cite{wang2024mmlu}, AI2ARC\cite{olea2024ai2arc}\\
        \midrule
        L1 & MOL\_Solubility & Text, SMILES & 109 & AqSolDB\cite{sorkun2019aqsoldb}, esol\cite{delaney2004esol} \\
         & MOL\_Freesolv & Text, SMILES & 43 & freesolv\cite{mobley2014freesolv} \\
         & MOL\_Syn & Text, SMILES & 176& USPTO-50K\cite{uspto} \\
         & MOL\_Resyn  & Text, SMILES &176 & USPTO-50K\cite{uspto}  \\
         & MOL\_Excited & Text, SMILES & 200 & QM8\cite{ramakrishnan2015qm8}  \\
         & MOL\_Thermo & Text, SMILES & 240 & QM9\cite{ruddigkeit2012qm9}  \\
         & MOL\_BBBP & Text, SMILES & 119 & BBBP\cite{martins2012bbbp}  \\
         & MOL\_HIV & Text, SMILES & 120 & MoleculeNet\cite{wu2018moleculenet}  \\
         & MOL\_HLGap & Text, SMILES & 251 & PCQM4Mv2\cite{hu2021ogblsc}, QM9\cite{ruddigkeit2012qm9}  \\
         & MOL\_TOX &  Text, SMILES & 76 & tox21\cite{wu2018moleculenet}  \\
        \midrule
        L2 &PROT\_Melt & Text, AA-Seq. & 133& Meltome\_atlas\cite{jarzab2020meltome}  \\
         & PROT\_Energy & Text, AA-Seq.  &101 & Tm262\cite{li2024prostage}, S669\cite{pancotti2022s669}  \\
         & PROT\_Fitness & Text, AA-Seq.  & 122& GB1\cite{wu2016gb1}, avGFP\cite{sarkisyan2016avgfp}  \\
         &PROT\_Mutation  & Text, AA-Seq.  &436 & ProteinGym\cite{notin2023proteingym}  \\
         &PROT\_Conserve  & Text, AA-Seq.  &104 & VESPA\cite{dallago2021flip}  \\
         & PROT\_Location &  Text, AA-Seq. & 107 & DeepLoc\cite{thumuluri2022deeploc2}  \\
         & PROT\_EC  &  Text, AA-Seq. & 196 & DeepFRI\cite{gligorijevic2021deepfri}  \\
         & PROT\_GO  &  Text, AA-Seq. & 258 & DeepFRI\cite{gligorijevic2021deepfri}  \\
         & PROT\_SSC  & Text, AA-Seq.  & 77 & CB513\cite{zhou2014cb513}, CASP12\cite{casp} \\
         & PROT\_Fold &  Text, AA-Seq. & 145 & PDB\cite{pdb} \\ 
        \midrule
        L3  & PPI\_Type  & Text, Two AA-Seqs. & 175 & SHS27k\cite{chen2019shs27k} \\
        & PPI\_Binary  & Text, Two AA-Seqs. & 159 & TAGPPI\cite{song2022tagppi} \\
        & DDI\_Interact & Text, Two SMILES & 170 & Drugbank\cite{knox2024drugbank} \\
        & PLI\_BA  & Text, AA-Seq., SMILES & 146 & SKEMPI\cite{jankauskaite2019skempi}, bindingdb\cite{liu2025bindingdb}, davis\cite{davis2011comprehensive}   \\
        & CI\_AbAg &  Text, Three AA-Seqs. & 90 & AbBiBench\cite{zhao2025abbibench} \\
        \bottomrule
    \end{tabularx}
    \label{tab:tab1}
\end{table}

\subsection{Comparative Performance of LLMs}

To demonstrate the utility and discriminability of our benchmark, 13 LLMs (Table \ref{tab:tab3}) are evaluated with detail prompt instruction (illustrated in Section \ref{sec:methods}), encompassing a range of architectural designs (dense and sparse, Transformer and Mamba), as well as both domain-specific and general-purpose models.
Table \ref{tab:tab2} presents a comprehensive ranking of the models.
The models are ranked from 1 (best) to 13 (worst) on each individual task, with the overall rank across all tasks summarized in the bottom row.

NVIDIA-Nemotron-Nano-9B-v2 emerges as the overall top performer with a composite rank of 1, demonstrating exceptional versatility across task categories.
It achieves 4 first-place rankings among others.
DeepSeek-V3.1 secures the second overall position, showing particularly strong performance on General\_Text task.
% While NVIDIA-Nemotron-Nano-9B-v2 demonstrates the most balanced performance, 
Moreover, several models exhibit domain-specific expertise.
Phi-4-14B achieves first-place rankings on two L3-level tasks, suggesting particular strength in interaction prediction tasks. 
% TxGemma-Chat-9B shows competitive performance on two tasks with ranking number 1, indicating the domain-specific robust performance despite a low rank in General\_Text.
% The ranking reveals substantial performance variations across models. 
However, NatureLM-8x7B generally falls below average across tasks, contributing to its overall lower standing.
% Similarly, Llama-3.1-8B-Instruct shows consistently poor performance, particularly on molecular and protein tasks.
% Notably, even top-performing models exhibit task-specific weaknesses, for instance, while Model E excels overall, it shows relatively weaker performance on PROT\_GO (rank 13) and PROT\_Fitness (rank 11).

% Certain tasks prove particularly discriminative among models. The PPI\_Type and PPI\_Binary tasks show clear performance stratification, with Models A, D, and E consistently outperforming others.
% In contrast, tasks like PLI\_BA show more compressed rankings, with multiple models (A, D, J, K) tied at rank 8, suggesting less discriminative power or potential task saturation. 
% Molecular property prediction tasks generally show strong performance from Models C, E, and M, while protein-centric tasks favor Models D and E.

The ranking demonstrates that no single model dominates across all tasks.
% , highlighting the continued need for task-specific model selection or ensemble approaches.
The strong performance of NVIDIA-Nemotron-Nano-9B-v2 suggests that the combination of mamba and attention architecture is more effective at capturing both global and local information simultaneously, offering inherent advantages for bio-molecular tasks, which typically involve long molecular SMILES strings and protein sequences as inputs.
The results of DeepSeek-V3.1 indicate that large-scale models with massive parameters, extensive data, advanced training strategies possess stronger knowledge retention capabilities compared to smaller models, thereby yielding broadly applicable representations.
The consistent under-performance of Llama-3.1-8B-Instruct and NatureLM-8x7B indicates their potential limitations for bio-molecular tasks.

% Variance across tasks within each model serves as an indicator of cross-domain stability. Models with narrower dispersion across both panels demonstrate stronger generalization and lower sensitivity to protocol modifications. Conversely, models with wide inter-task spread indicate domain-specific strengths but limited robustness.

% Overall, our findings demonstrate that while LLMs hold immense potential for biomolecular analysis, targeted improvements in robustness, reasoning, and domain adaptation are necessary. This benchmark and analysis provide a foundation for future research toward more reliable scientific LLMs.

\begin{table}[h]
    \centering
    \caption{\textbf{Overview of evaluated models.}
    The table lists 13 language models evaluated, models vary in the number of parameters (\#Params), open-source availability (Open), whether the model has undergone domain-specific fine-tuning (Domain-FT), and the underlying architectural type (Architecture).
    MoE denotes Mixture-of-Experts architecture, while D.T. refers to Dense Transformer.
    Parameter counts marked with "/" indicate undisclosed model sizes.
    % The top section includes large-scale models (>50B parameters), followed by domain-fine-tuned models (middle section), and general-purpose instruction-tuned models (bottom section).
    }
    \label{tab:tab3}
    % \begin{tabular}{lcccc}
    % \begin{tabularx}{\textwidth}{XXXXX}
    \begin{tabularx}{\textwidth}{p{7.0cm}p{1.8cm}p{1.2cm}p{2.2cm}p{2.0cm}}
        \toprule
        Model & \#Params & Open & Domain-FT & Architecture\\
        \midrule
        DeepSeek-v3.1\cite{deepseekv3.1} & 685B & \ding{51} & \ding{55} & MoE \\
        Qwen3-235b-a22b-2507\cite{bai2023qwen} & 235B  & \ding{51} & \ding{55} & MoE \\
        GPT-5-mini\cite{gpt5mini} & /  & \ding{55} & \ding{55} &  MoE \\
        \midrule
        NatureLM-8x7B\cite{xia2025naturelm} & 56B & \ding{51} & \ding{51} & MoE \\
        TxGemma-Chat-9B\cite{wang2025txgemma} & 9B & \ding{51} & \ding{51} & D.T. \\
        \midrule
        Phi-4-14B\cite{abdin2024phi4} & 14B & \ding{51} & \ding{55} & D.T.  \\
        GPT-oss-20B\cite{agarwal2025gptoss} & 20B & \ding{51}  & \ding{55} & D.T.  \\
        Gemma-3-12B\cite{team2024gemma} & 12B & \ding{51}  & \ding{55} & D.T.  \\
        Qwen-3-14B\cite{bai2023qwen} & 14B & \ding{51} & \ding{55} & D.T.  \\
        NVIDIA-Nemotron-Nano-9B-v2\cite{basant2025nemotron} & 9B & \ding{51} & \ding{55}  & Mamba  \\
        Mistral-Nemo-Instruct-2407\cite{mistral2024inst} & 12B & \ding{51} & \ding{55} & D.T.  \\
        Llama-3.1-8B-Instruct\cite{touvron2023llama} & 8B & \ding{51} & \ding{55}  & D.T.  \\
        DeepSeek-R1-Distill-Qwen-14B\cite{liu2024deepseek} & 14B & \ding{51} & \ding{55} & D.T.  \\
        \bottomrule
    \end{tabularx}
\end{table}

\begin{table}[h]
    \centering
    \caption{\textbf{Model rankings on each benchmark tasks.} The overall rank is obtained by calculating the ranking of all models based on the average of each model's rankings across all tasks. A: Phi-4-14B, B: GPT-oss-20B, C: Gemma-3-12B, D: Qwen-3-14B, E: NVIDIA-Nemotron-Nano-9B-v2, F: Mistral-Nemo-Instruct-2407, G: Llama-3.1-8B-Instruct, H: DeepSeek-R1-Distill-Qwen-14B, I: TxGemma-Chat-9B, J: NatureLM-8x7B, K: Qwen3-235b-a22b-2507, L: GPT-5-mini, M: DeepSeek-v3.1.}
    % \resizebox{\textwidth}{!}
    % {
        % \begin{tabularx}{l*{13}{c}}
        \begin{tabularx}{\textwidth}{lXXXXXXXXXXXXX}
        \toprule
        \textbf{Tasks} & 
        \textbf{A} & 
        \textbf{B} & 
        \textbf{C} & 
        \textbf{D} & 
        \textbf{E} & 
        \textbf{F} & 
        \textbf{G} & 
        \textbf{H} & 
        \textbf{I} & 
        \textbf{J} & 
        \textbf{K} & 
        \textbf{L} & 
        \textbf{M} \\
        \midrule
        General\_Text          & 3  & 11 & 8  & 5  & 7  & 10 & 9  & 1  & 12 & 13 & 4  & 6  & 2  \\
        PROT\_EC               & 12 & 13 & 8  & 1  & 3  & 4  & 5  & 7  & 10 & 11 & 6  & 9  & 2  \\
        PROT\_Location         & 11 & 13 & 7  & 1  & 5  & 2  & 4  & 10 & 6  & 12 & 3  & 9  & 8  \\
        PPI\_Type              & 1  & 12 & 9  & 2  & 3  & 4  & 8  & 10 & 5  & 13 & 7  & 11 & 6  \\
        PPI\_Binary            & 1  & 13 & 6  & 8  & 4  & 2  & 11 & 10 & 7  & 12 & 5  & 9  & 3  \\
        MOL\_TOX               & 9  & 12 & 8  & 3  & 1  & 10 & 7  & 11 & 2  & 13 & 5  & 4  & 6  \\
        MOL\_HIV               & 8  & 13 & 2  & 9  & 3  & 4  & 6  & 11 & 10 & 12 & 1  & 7  & 5  \\
        MOL\_BBBP              & 1  & 13 & 7  & 3  & 5  & 8  & 10 & 11 & 6  & 12 & 9  & 4  & 2  \\
        PLI\_BA                & 8  & 7  & 6  & 8  & 2  & 1  & 13 & 4  & 12 & 8  & 8  & 5  & 3  \\
        MOL\_Thermo            & 7  & 11 & 8  & 9  & 3  & 2  & 13 & 1  & 12 & 10 & 4  & 5  & 6  \\
        MOL\_Excited           & 7  & 9  & 1  & 4  & 5  & 11 & 13 & 3  & 12 & 6  & 8  & 10 & 2  \\
        PROT\_Energy           & 3  & 8  & 6  & 2  & 1  & 11 & 13 & 4  & 12 & 10 & 9  & 5  & 7  \\
        MOL\_Freesolv          & 6  & 8  & 2  & 3  & 4  & 10 & 13 & 1  & 12 & 7  & 11 & 9  & 5  \\
        MOL\_HLGap             & 2  & 10 & 1  & 3  & 4  & 11 & 13 & 5  & 12 & 9  & 6  & 7  & 8  \\
        PROT\_Mutation         & 7  & 4  & 8  & 6  & 3  & 10 & 13 & 1  & 12 & 11 & 9  & 2  & 5  \\
        MOL\_Solubility        & 1  & 8  & 6  & 9  & 2  & 11 & 13 & 3  & 12 & 10 & 5  & 7  & 4  \\
        PROT\_Fitness          & 4  & 5  & 6  & 7  & 2  & 11 & 13 & 1  & 12 & 10 & 9  & 3  & 8  \\
        PROT\_Melt             & 5  & 4  & 7  & 9  & 1  & 11 & 13 & 2  & 12 & 10 & 6  & 3  & 8  \\
        CI\_AbAg               & 1  & 10 & 9  & 6  & 2  & 11 & 13 & 3  & 12 & 8  & 5  & 7  & 4  \\
        PROT\_GO               & 10 & 11 & 6  & 3  & 13 & 7  & 9  & 2  & 1  & 12 & 5  & 8  & 4  \\
        DDI\_Interact          & 7  & 8  & 9  & 5  & 1  & 13 & 10 & 11 & 2  & 12 & 3  & 6  & 4  \\
        MOL\_Syn               & 7  & 13 & 2  & 5  & 10 & 6  & 8  & 11 & 1  & 12 & 3  & 9  & 4  \\
        MOL\_Resyn             & 5  & 10 & 2  & 4  & 8  & 1  & 7  & 12 & 9  & 11 & 3  & 13 & 6  \\
        \midrule
        \textbf{Overall Rank} & \textbf{4} & \textbf{11} & \textbf{5} & \textbf{3} & \textcolor{red}{\underline{\underline{\textbf{1}}}} & \textbf{9} & \textbf{12} & \textbf{7} & \textbf{10} & \textbf{13} & \textbf{6} & \textbf{8} & \underline{\textbf{2}} \\
        \bottomrule
        \end{tabularx} %
    % }
    \label{tab:tab2} 
\end{table}

\subsubsection{Classification tasks}

\begin{figure}[h]   %[htbp]
    \centering
    \includegraphics[clip, trim=0cm 3cm 0cm 0cm, width=1.0\textwidth]{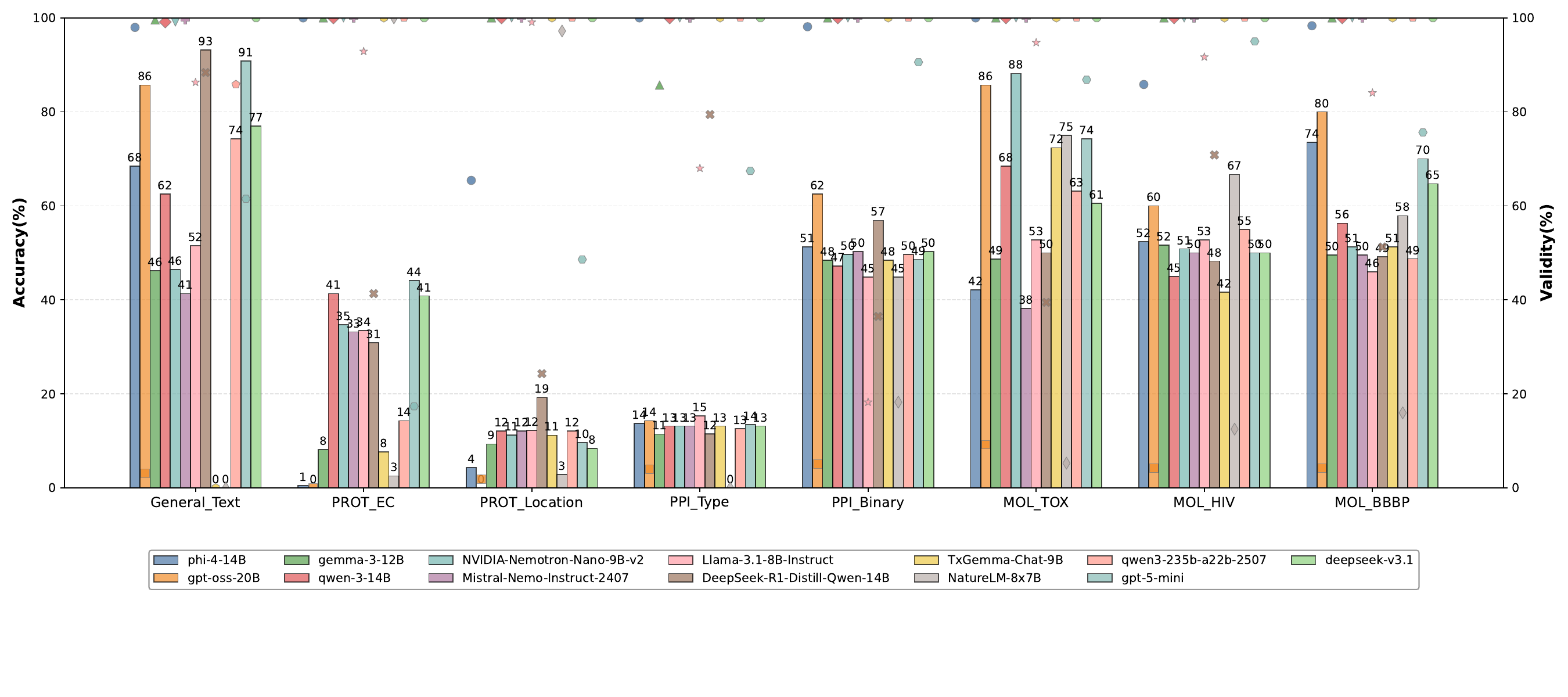}
    \caption{\textbf{Performance of 13 LLMs among 8 classification tasks.}
    The prediction accuracy are displayed with bars based on the left axis.
    The markers represent model output validity based on the right axis.
    Larger-parameter models rank among the top performers in overall accuracy.
    % Lower validity indicates that syntactic compliance remains a non-trivial factor in automated evaluation pipelines.
    }
    \label{fig:fig2}
\end{figure}

Firstly, we analyze model performance across 8 classification tasks, evaluated in terms of prediction accuracy and output validity.
% The tasks include General\_Text, PROT\_EC, PROT\_Location, PPI\_Type, PPI\_Binary, MOL\_TOX, MOL\_HIV, and MOL\_BBBP. 
% Together, these tasks span general molecular understanding, protein functional annotation, protein–protein interaction prediction, and small molecule pharmacological property assessment.
% Overall performance trends
As Figure \ref{fig:fig2} shows, models achieve higher performance on text-only (General\_Text) or easy classification tasks(MOL\_BBBP and MOL\_Tox), whereas performance declines for structurally complexes or biologically nuanced tasks such as protein–protein interaction type prediction.
Specifically, due to the deep fine-tuning process of NatureLM-8x7B, its accuracy on some tasks is zero as the fine-tuning process did not learn it.
Large-parameter models, including Qwen3-235b-a22b-2507 and DeepSeek-v3.1, consistently rank among the top performers in overall accuracy. 
% These models demonstrate comparatively stable performance across heterogeneous task categories, suggesting improved cross-domain generalization. 
Medium-scale models such as Qwen-3-14B and Gemma-3-12B exhibit greater fluctuations and reduced robustness.

As for output validity (indicated with markers on Figure \ref{fig:fig2}), we measure whether model outputs conform to the required answer format and produce parsable predictions.
Validity generally remains high for the majority of models, while certain models such as GPT-oss-20B demonstrate reduced validity in complicated tasks.
% Lower validity often correlates with tasks that demand precise formatting, indicating that syntactic compliance remains a non-trivial factor in automated evaluation pipelines.
Notably, some high-accuracy models such as Phi-4-14B occasionally show reduced validity in specific categories, suggesting that raw predictive capability does not necessarily guarantee strict adherence to output constraints. 
This distinction underscores the importance of jointly evaluating semantic correctness and formatting reliability in LLM benchmarking.

% Cross-scale generalization 
% When aggregating across all tasks, larger and more recent architectures demonstrate improved cross-scale generalization, maintaining comparatively stable performance from general textual reasoning to protein-level and molecular-level prediction tasks.
% In contrast, smaller or distilled models show sharper performance declines as task complexity increases. 
% This pattern suggests that parameter scale and architectural refinement contribute to improved integration of heterogeneous bio-molecular representations.

% Performance dispersion
% Inter-task variance is more pronounced than inter-model variance within the same scale category. This indicates that task formulation and domain complexity exert a stronger influence on performance than architectural differences among models of similar size. In particular, protein interaction subtype classification and toxicity-related tasks contribute disproportionately to overall performance variance.
\subsubsection{Regression tasks}

\begin{figure}[h]   %[htbp]
    \centering
    \includegraphics[clip, trim=0cm 2cm 0cm 0cm, width=1.0\textwidth]{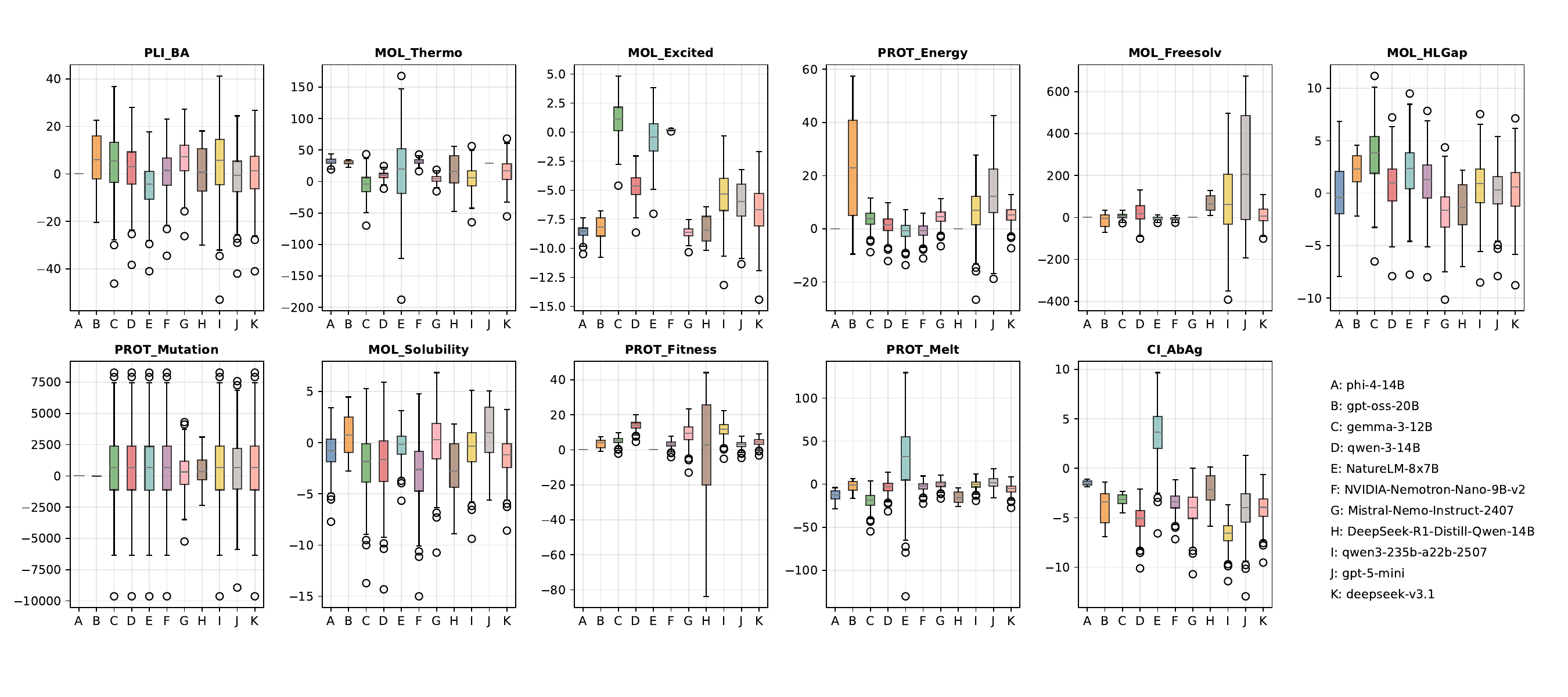}
    \caption{\textbf{The distributions of prediction results on eleven regression tasks.}
    By calculating the difference between each model's predictions and the ground-truth labels across different samples for each task, the mean and variance are computed, from which the boxplot in each subplot is generated.
    }
    \label{fig:fig3}
\end{figure}

Figure \ref{fig:fig3} presents model performance across eleven bio-molecular regression tasks spanning molecular thermodynamics, quantum properties and protein mutational effects.
% , protein stability, fitness prediction, binding affinity estimation, and antibody–antigen interaction modeling.
% Models A–K correspond to increasing scale and architectural diversity, ranging from mid-sized instruction-tuned models to large-scale foundation models.
Overall performance distribution across all tasks exhibits substantial variability.
% Error magnitudes differ significantly across task types, reflecting heterogeneity in target scales and physicochemical complexity. 
Tasks involving protein-level stability (PROT\_Mutation) and thermodynamics (MOL\_Thermo) exhibit wider numerical ranges compared to quantum chemical tasks (MOL\_Excited and MOL\_HLGap).
NVIDIA-Nemotron-Nano-9B-v2 and larger-scale models including DeepSeek-v3.1 generally demonstrate lower absolute error magnitudes and tighter dispersion across most tasks.
Notably, performance improvements with scale are not uniform across all tasks. 
While L2-level tasks generally benefits from increased parameter count, L1-level tasks show diminishing returns, indicating that scaling alone may not resolve biologically complexes inference challenges.
% NVIDIA-Nemotron-Nano-9B-v2 emerges as the most versatile regressor, particularly excelling on tasks involving molecular and protein property prediction.

After parsing, Llama-3.1-8B-Instruct and TxGemma-Chat-9B models are unable to produce prediction results on any tasks, and are therefore not shown in Figure \ref{fig:fig3}.
All models perform poorly on two extremely difficult tasks (MOL\_Thermo and PROT\_Mutation), indicating that future work should investigate ensemble methods to better handle the challenging long-sequence regression tasks prevalent in bio-molecular domains.
% The extreme difficulty of  across all models suggests the need for task-specific architectural or model training adaptations. 

\subsubsection{Generation tasks}

\begin{figure}[h]   %[htbp]
    \centering
    \includegraphics[clip, trim=0cm 0cm 0cm 0cm, width=1.0\textwidth]{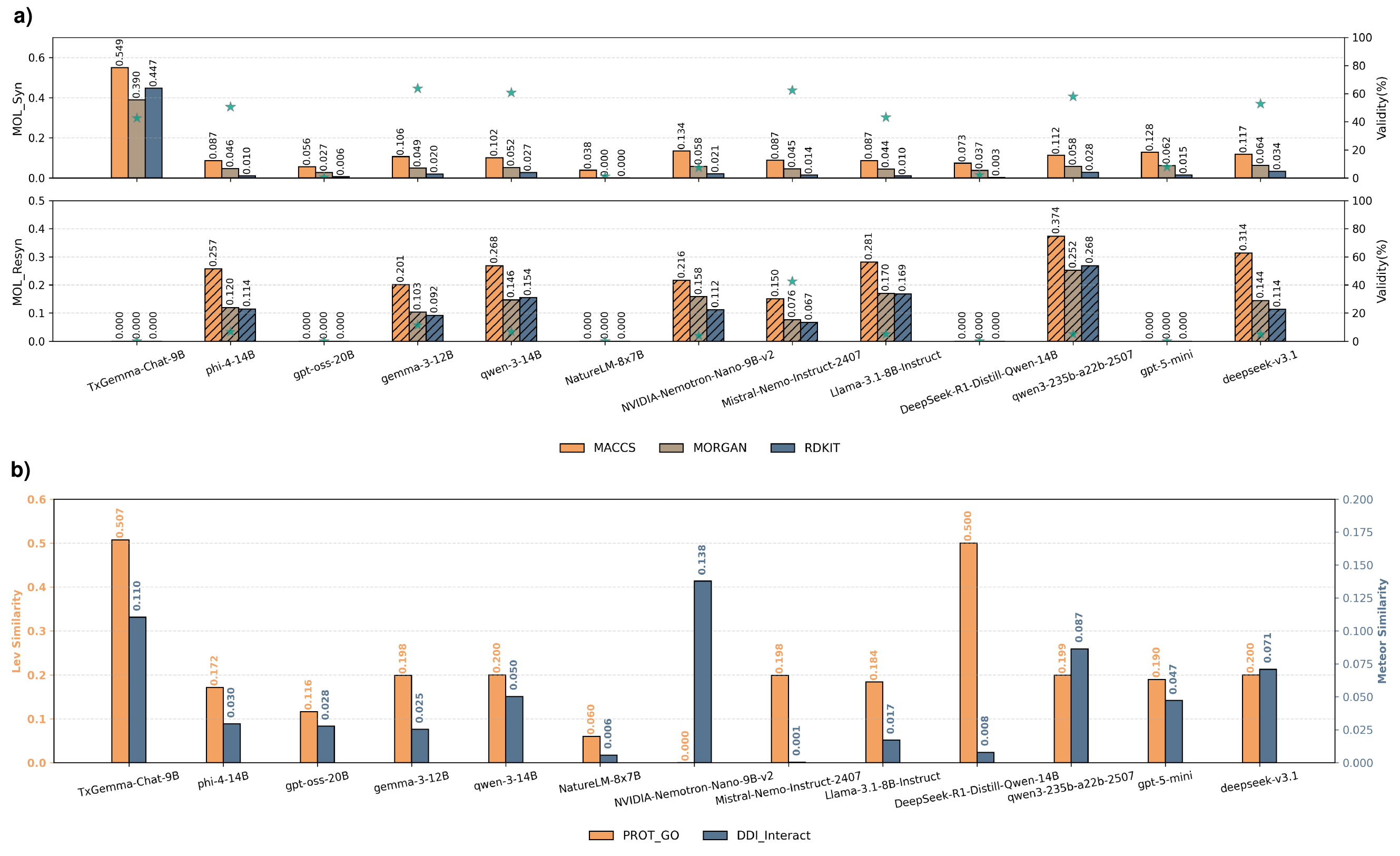}
    \caption{\textbf{Model performance on generation tasks.}
    Higher values indicate better model performance.
    \textbf{a).} For MOL\_Syn and MOL\_Resyn tasks, fingerprint similarities of the prediction results were calculated using three methods: MACCS, MORGAN, and RDKIT. 
    These values are displayed as bars corresponding to the left y-axis.
    The markers that represent the validity of the output results correspond to right y-axis. 
    \textbf{b).} The results for PROT\_GO and DDI\_Interact were evaluated with Levenshtein similarity and Meteor similarity metric, respectively.
    % represented by orange bars. The results for DDI\_Interact are evaluated with  and corresponded to blue bars.
    % Larger similarity values indicate better performance.  
    }
    \label{fig:fig4}
\end{figure}

There are 7 generation tasks in our benchmark, Figure \ref{fig:fig4} presents a comparative analysis of model behavior on 4 tasks.
% under two complementary evaluation settings, shown in panels (a) and (b). Together, these panels illustrate differences in performance stability, scaling behavior, and cross task consistency under distinct experimental configurations.
% The top panel summarizes the quantitative performance on MOL\_Syn and MOL\_Resyn tasks.
% The results of PROT\_GO and DDI\_Interact are shown in the bottom panel.
As for other 3 extremely hard tasks (PROT\_Conserve, PROT\_Location and PROT\_Fold), all models are unable to complete, so the results are not shown.
% Across models, performance exhibits systematic variation consistent with model scale and architectural capacity.
As for MOL\_Syn task, TxGemma-Chat-9B demonstrates significant advantage over other models, based on the fingerprint similarity results using MACCS, MORGAN and RDKIT calculation methods.
The results of other models show relatively minor differences. 
Furthermore, the fingerprint result distributions obtained from 3 calculation methods are consistent.
Similarly, on PROT\_GO and DDI\_Interact tasks (panel b), TxGemma-Chat-9B demonstrates leading performance, highlighting the advantage of this model's exceptional capabilities on generative tasks after fine-tuning with domain knowledge.

On the MOL\_Resyn task, the overall performance of all models is mediocre.
Among them, the Qwen3-series models achieves the best results. 
Five models including NatureLM-8x7B and DeepSeek-R1-Distill-Qwen-14B, are completely unable to generate valid molecules for the task. 
Notably, DeepSeek-R1-Distill-Qwen-14B, which was fine-tuned with chain-of-thought (CoT) based on Qwen-3-14B, exhibits weakened generative capabilities compared to its baseline model.
It indicates that the CoT fine-tuning approach does not have a significant improvement effect on specific tasks in the bio-molecular domain.

\subsection{Tool Capability Comparison}

\begin{figure}[h]   %[htbp]
    \centering
    \includegraphics[width=1.0\textwidth]{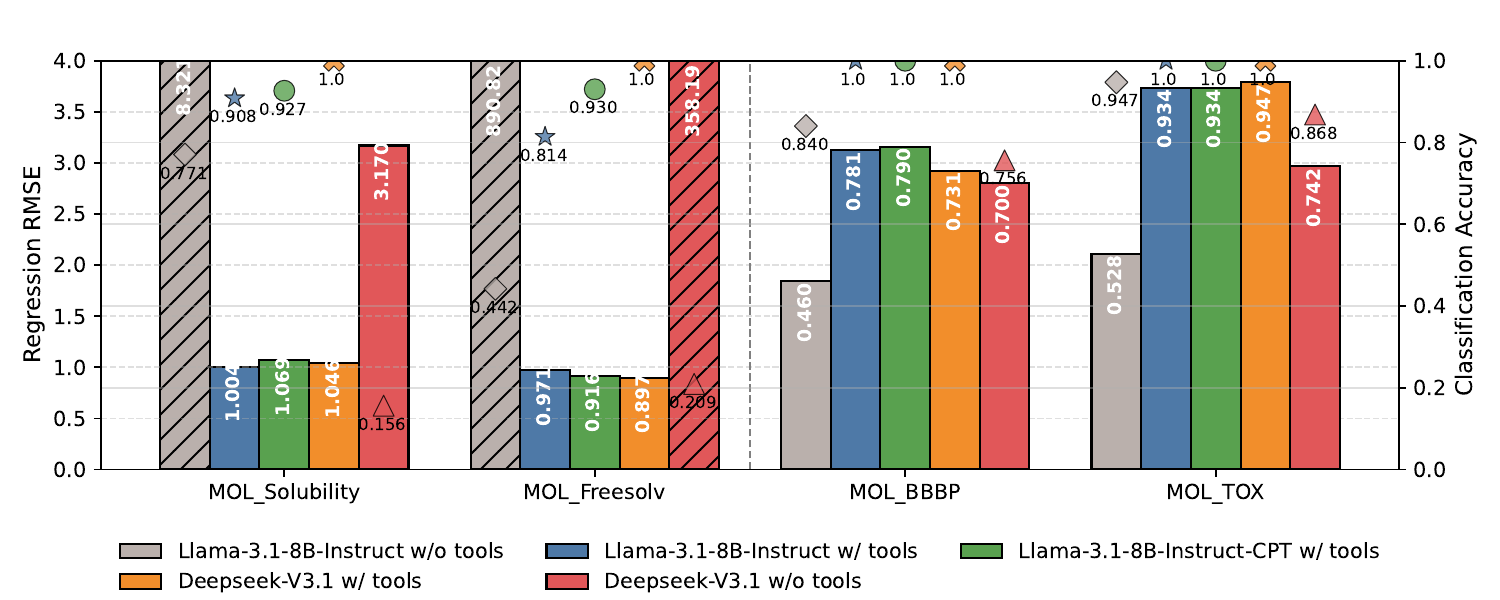}
    \caption{\textbf{LLM performance with and without tool integration across 4 representative tasks.}
    (left) For regression-based tasks (MOL\_Solubility and MOL\_Freesolv), performance is evaluated using RMSE, where lower value bars indicate better performance.
    The dashed bars indicate values exceeding the coordinate range, and the actual value is displayed with white numerical text.
    (right) For classification tasks, performance is measured using accuracy.
    The markers represent model output validity.}
    \label{fig:fig5}
\end{figure}

We compare benchmark performance between single LLMs using direct prompt and agentic workflows with tool integration.
Figure \ref{fig:fig5} exhibits the performance of two different backbone models (DeepSeek-v3.1 and Llama-3.1-8B) on two regression and two classification tasks.
Firstly, across all tasks and regardless of which model, tool-enabled configurations exhibit greater numerical stability and improved overall performance.
The most striking effect is observed in regression tasks, where disabling tools leads to extreme error values, suggesting that unconstrained language generation is insufficient for precise quantitative estimation.
Classification tasks show smaller but consistent performance drops without tools, implying that symbolic reasoning alone partially supports binary decision-making but lacks reliability.

Secondly, we compare the effect of continual pretraining (CPT).
Llama-3.1-8B-Instruct-CPT is trained with sufficient bio-molecular knowledge.
CPT introduces slight improvement in both regression and classification tasks.
This indicates that domain-specific CPT enhances the model’s ability to interact effectively with structured chemical inputs and tool outputs.
Finally, comparing the results of models of different parameter scales, it is counter-intuitive that although DeepSeek-V3.1 has over 50 times more parameters than LLama-3.1-8B, their overall performance is similar, outperforming each other in two tasks respectively.
% Overall, the experimental results clearly indicate that the integrated tool has a significant additive effect, while CPT and increasing model size have slight impact.

% Following the breakthrough of AlexNet \cite{krizhevsky2012imagenet} in the ImageNet competition, a series of deep neural networks have been proposed, including VGG \cite{simonyan2014very}, GoogLeNet \cite{szegedy2015going}, and ResNet \cite{he2016deep}.

\subsection{Ablation study of experiment settings}

\begin{figure}[h]   %[htbp]
    \centering
    \includegraphics[width=1.0\textwidth]{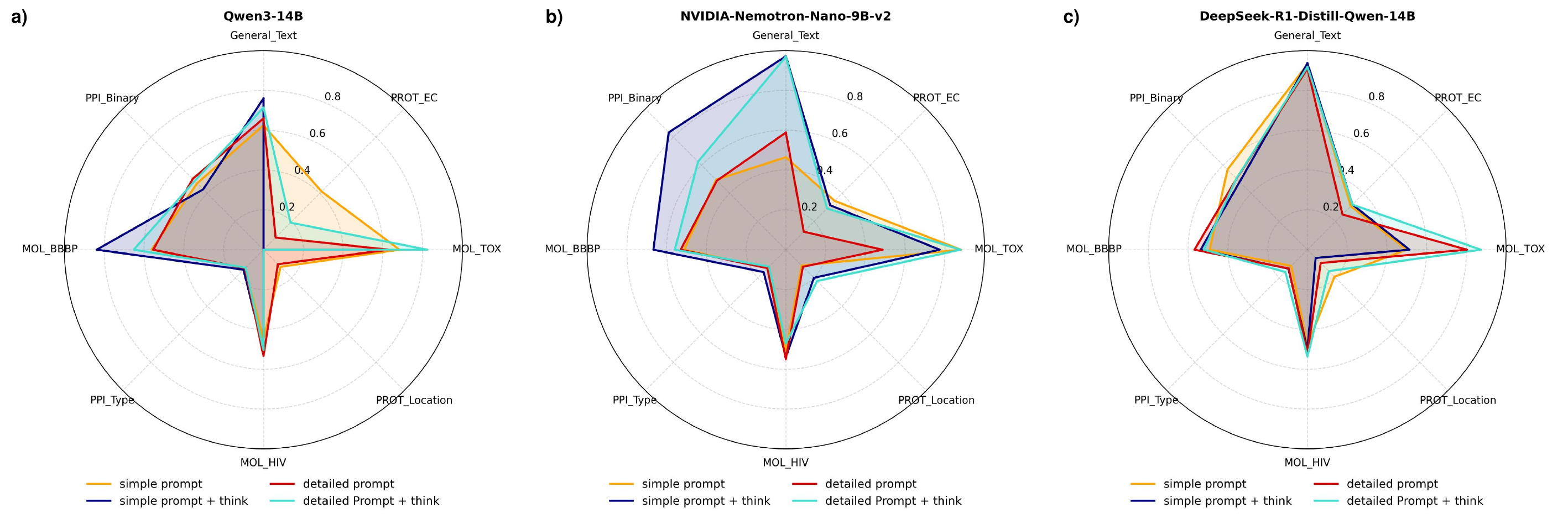}
    \caption{\textbf{Model behavior under different input prompt complexity and think mode.} 
    Performance is evaluated with accuracy.
    Detailed prompt with more domain knowledge guidance generally exhibits better performance, and model performance under think mode varies depending on the task.
    NVIDIA-Nemotron-Nano-9B-v2 shows higher accuracy and sensitivity to different experiment settings.}
    \label{fig:fig6}
\end{figure}

We also examine model sensitivity to different prompts and think mode as minor variations in phrasing, formatting, or instruction style may lead to performance fluctuations.
%, with drops of up to several percentage points in accuracy on identical tasks. This underscores the brittleness of LLMs in scientific contexts, where precise query formulation is essential, and motivates the need for prompt-robust evaluation protocols in future benchmarks.
Qwen3-14B, NVIDIA-Nemotron-Nano-9B-v2 and DeepSeek-R1-Distill-Qwen-14B are selected because they support think mode.
% Results are demonstrated in Figure \ref{fig:fig6}, which highlights both quantitative performance shifts and structural differences in cross-task and cross-model behavior. 
% presents a two-panel analysis of model behavior under an alternative experimental configuration.
% Panels (a) and (b) collectively examine how performance distribution and robustness change under modified evaluation conditions. The figure 
% Panel (a): Performance magnitude and distribution
Simple and detailed prompt template are described in Section \ref{sec:methods}.
Overall, NVIDIA-Nemotron-Nano-9B-v2 (Figure \ref{fig:fig6}) shows stronger sensitivity and variability to different think mode and input prompt than Qwen3-14B.
And model performance on hard tasks such as PPI\_Binary can be improved by think process
On most tasks, DeepSeek-R1-Distill-Qwen-14B's sensitivity to prompt is low, with only significant differences observed on MOL\_TOX task.
% So when using this model, detailed knowledge guidance should be provided as much as possible.
% As for think mode, the easy tasksthe results is also differentiated. 
% For LLMs that have been exposed to tasks during pretraining stage, the model can improve performance through the think process.
These results indicate that when using LLM to complete domain-specific tasks, more domain knowledge guidance should be given, and whether to use think mode needs to be flexibly adjusted according to the difficulty of tasks.
% As for Gemma-3-12b-it, the result exhibits subtle performance differences in different experiment settings for all tasks.
% When using this model to complete tasks related to bio-molecules, prompt construction is more open and there is no need to use think mode, which can improve execution speed.

\subsection{LLM reasoning ability in scientific problem}
To investigate how LLMs solve scientific problem, we analyze their CoT reasoning processes.
Beyond final-answer accuracy, this analysis focuses on whether intermediate reasoning reflects a principled understanding of biomolecular mechanisms.
% or instead relies on heuristic pattern matching.
A representative example arises from molecular solvation free energy prediction task as follows.
% A model produces the correct prediction (“binding affinity decreases”), accompanied by the following CoT-style explanation:
At first glance, this reasoning appears chemically sound. 
However, the reasoning processing contains some contradictions and chemical inconsistencies: If the molecule is fused and contains two ketones, it is more likely a naphthoquinone derivative or a dione, but not similar to acetanilide.
% This example highlights a broader pattern: the model arrives at the correct answer but with wrong mechanistic reasons.
% Its explanation is constructed from generally valid biochemical principles, yet it is not grounded in the specific structural context of the problem.
Similar behavior is observed across task types and models.
In generative tasks such as forward synthesis, models produce fluent and domain-relevant rationales, but lack a verifiable linkage to the governing chemical processes.
These observations indicate that LLMs solve scientific problems primarily through a combination of pattern recognition, abstraction, and heuristic reasoning, rather than through explicit modeling of underlying mechanisms.

% second question output
% \needspace{10cm}
\begin{mdframed}[frametitle={Output example of DeepSeek-R1-Distill-Qwen-14B under think mode:}, backgroundcolor=white, leftmargin=5pt, rightmargin=5pt, skipabove=15pt, linewidth=1pt]
  Okay, so I need to predict the free energy of a molecule given its SMILES notation. The SMILES provided is c1ccc2c(c1)C(=O)c3cccc(c3C2=O)N. Hmm, let me try to figure this out \textbf{step by step}.

\textbf{First}, [... Omitted for brevity ...]

\textbf{Putting this together}, the molecule might have \textbf{two benzene rings connected in a fused manner}, with two ketone groups and an amine substituent. This structure could be \textbf{similar to acetanilide} or something more complex, perhaps a beta-lactam or another bicyclic compound.

\end{mdframed}

\section{Discussion}
\label{sec:discussion}
% key takeaways: 2 3 4 7
% 1. cross scale benchmark 很好的评估了不同模型在生物分子领域的能力
% 2. 模型架构对模型性能的影响最为显著 mamba的长序列信息学习能力 成就了最优表现
% 3. 模型规模 对模型在生物分子任务能力上 起次要作用
% 4. 经过深度SFT的模型 变成了书呆子 泛化能力缺失
% 5. 对于difficult tasks 目前的模型均无法完成 这也是未来大模型优化的方向
% 6. agentic workflow通过tool的集成 配合LLM的任务解析能力 大幅提升在benchmark上的表现
% 7. 其他参数配置 如是否使用think模型、是否CoT模型、prompt复杂度 对模型性能的影响 在不同task上的表现各不相同 实际推理时 应该视任务情况灵活调整

The cross-scale bio-molecular benchmark introduced in this work provides a comprehensive evaluation framework for assessing large language models on biologically-relevant tasks spanning multiple scales of molecular complexity. 
Through systematic evaluation of 13 state-of-the-art models, we have elucidated several important insights regarding model capabilities, architectural influence, and the value of tool integration in the bio-molecular domain.

Our ranking analysis reveals that NVIDIA-Nemotron-Nano-9B-v2, which employs a hybrid mamba-attention architecture, achieves the best overall performance across BioMol-LLM-Bench. 
% This finding carries significant implications for model design in scientific domains. 
This suggests that hybrid architecture appears particularly well-suited for processing the long-range dependencies inherent in bio-molecular sequences.
%, such as extended SMILES strings and protein FASTA sequences. 
%his suggests that purely transformer-based architectures, while dominant in general language tasks, may benefit from hybridization with state-space models for long-range dependent tasks.
% \item The strong performance of DeepSeek-v3.1, ranking second overall, demonstrates that parameter scale and training data diversity remain crucial factors for scientific reasoning. 
Poor performance across all models on 2 challenging tasks (MOL\_Thermo and PROT\_Mutation) highlights the difficulty of long-sequence regression in bio-molecular applications and motivates exploration of model optimization.
Models that underwent extensive domain-specific fine-tuning, such as TxGemma-Chat-9B, demonstrate remarkable proficiency on a small portion of tasks yet exhibit significant degradation on general text understanding.
This mechanism reflects a degradation in generalization capability resulting from SFT.
Through the analysis of reasoning process, we found that LLMs may arrive at the correct answer but with wrong mechanistic reasons.
% \item The model sensitivity to prompt variations, think mode and CoT fine-tuning, underscores the importance of careful model selection for bio-molecular applications.

% Overall, the substantial variability observed across task categories reveals important distinctions in model capabilities. 
Overall, the consistently high performance on lower-level tasks (L0-L1 levels) suggests that current LLMs have developed robust representations for small molecular language when expressed in natural language format.
However, the marked performance degradation on L3-level tasks involving protein-protein interactions indicates a fundamental limitation in current models' ability to reason about relationships between molecules.
% Perhaps the most practically significant finding is the dramatic performance improvement conferred by tool integration. 
The observation that tool-enabled configurations consistently outperform direct prompting, supports a hybrid intelligence paradigm where LLMs serve as reasoning orchestrators while specialized tools handle quantitative computations.
% The extreme error values observed when tools were disabled for regression tasks suggest that language models, despite their parametric knowledge, cannot reliably perform precise quantitative estimation without computational assistance. 
In conclusion, the cross-scale bio-molecular benchmark provides both a standardized evaluation methodology and empirical insights that are promosing to guide future development of language model design for bio-molecular discovery.
% As the field advances toward agentic workflows, the integration of language models with domain-specific computational tools, guided by comprehensive benchmarks that capture the multi-scale nature of bio-molecular problems, will be essential for realizing the transformative potential of AI in the life sciences.

\section{Methods}

\label{sec:methods}

\subsection{Benchmark Construction}

% The construction of the benchmark involved sourcing data from established open-source datasets, followed by systematic preprocessing, task definition, prompt standardization, and evaluation metric specification.
% All processing steps were implemented using Python 3.10 with libraries including RDKit (version 2023.03.3), Pandas, and scikit-learn.
% Metrics were computed using scikit-learn and RDKit utilities. For all tasks, results were reported with 95\% confidence intervals where applicable, derived from bootstrapping or cross-validation splits.

% ========================================================== %
% write in Method.
We proposes a comprehensive benchmark dataset specifically tailored for evaluating the capabilities of LLMs in molecular sciences. 
The source dataset is carefully derived from multiple high-quality sources to ensure diversity and reliability. 
Moreover, tool calling API and unified evaluation pipeline are integrated to facilitate efficient model comparison.
% All processing steps were implemented using Python 3.12 with libraries such as RDKit, Pandas and scikit-learn.

\textbf{1. Diverse Raw Open-Source Data Sources. }
The foundation of the benchmark draws from a variety of raw open-source data, combining domain-specific benchmarks (e.g., MoleculeNet\cite{wu2018moleculenet} suites including QM9 for quantum properties and Tox21 for toxicity) with general semantic understanding benchmarks (e.g., biomolecule-related questions in MMLU-Pro and AI2ARC).
Detailed descriptions of tasks are given in Table \ref{tab:tab1} and Supplementary Table \ref{app:descript}.
This integration ensures the dataset to capture both specialized molecular tasks and broader chemical knowledge.

\textbf{2. Standardized Data Processing Workflow. }
To maintain consistency and quality, a rigorous standardized workflow is applied.
\begin{itemize}
    \item \textbf{Filtering.} Removes invalid, incomplete, or noisy entries (e.g., invalid SMILES strings or duplicate molecules) with RDKIT software and DeepSeek-v3.1 model. 
    % Prompt for DeepSeek-v3.1 to filter bio-molecular samples is given below.  shadow=true, shadowsize=1pt,
        \begin{mdframed}[frametitle={Prompt for sample filtering:}
                , backgroundcolor=white, leftmargin=5pt, rightmargin=5pt, linewidth=1pt]          
            You are an expert in life science.
            
            Please carefully analyze the following question and judge that whether (Yes or No) the question related to small molecule and protein science, and give the confidence score (between 0 and 1) about your answer.
            
            Question: \{input\_text\}       
            
            Please answer in exactly the following format:
            
            Answer: 
            
            Confidence:
        \end{mdframed}
   \item \textbf{Cleaning.} Normalizes representation of molecules to canonical SMILES format and resolves inconsistencies in labels.
   \item \textbf{Merging.} Integrates data from multiple sources while avoiding overlaps through similarity checks.
   \item \textbf{Length casting.} Natural language text with excessive byte counts (more than 1000) or protein sequence longer than 500 characters were filtered out to reduce computational complexity.
   \item \textbf{Sampling.} Employs stratified sampling to balance molecular complexity and property distributions.  
   \item \textbf{Metadata supplement.} For complicated tasks such as protein ligand interaction prediction, additional protein and molecule structure information are provided.
\end{itemize}
   
% This pipeline yields a clean, high-fidelity dataset suitable for instruction tuning and benchmarking LLMs.

\textbf{3. Coverage of Diverse Tasks Across Molecular Scales. }
% appendix , task description TODO
The benchmark encompasses 4 levels of tasks according to input complexity.
\begin{itemize}
    \item \textbf{L0: General bio-molecular knowledge understanding.} Input data type of questions in this level only contains natural language.  
    \item \textbf{L1: Small-molecule level tasks.} Focusing on small molecule property prediction including chemical and physiological property. Both natural language and SMILES representation of molecules are provided.
    \item \textbf{L2: Large-molecule level task.} Focusing on tasks related to protein function and structure. Input data modalities include natural language and protein sequence.
    \item \textbf{L3: Multiple-molecules level task.} Incorporating challenging real-world tasks, such as antibody-antigen binding prediction, with more input data types including molecule SMILES, protein sequence, and natural language.  
\end{itemize}

Processed datasets for each task were stored in standard JSON format, with columns including canonical SMILES, protein sequence, task-specific labels, and metadata.

% This multi-scale coverage tests LLMs' ability to handle hierarchical chemical reasoning, from fundamental principles to practical applications in drug discovery and materials science.

\textbf{4. Automated Result Parsing and Evaluation Pipeline. }
To enable fair and reproducible comparison of LLMs, we proposed automatic output parsing protocol, through which the extracted type-specific answers of different models were obtained, and the performance of the model is further evaluated through standard task-specific evaluation metrics.
\begin{itemize}
    \item Classification tasks contain MOL\_TOX, MOL\_HIV, MOL\_BBBP, PPI\_Binary, PROT\_EC, PROT\_Location, General\_Text and PPI\_Type. Model performance is evaluated through accuracy and answer-type validity.  
   \item Regression tasks contain MOL\_HLGap, MOL\_Thermo, MOL\_Excited, CI\_AbAg, PLI\_BA, PROT\_Mutation, PROT\_Fitness, PROT\_Energy, PROT\_Melt, MOL\_Freesolv and MOL\_Solubility. Model performance is evaluated through RMSE, MEAN, STD and answer-type validity.  
   \item Generation tasks contain MOL\_Resyn, MOL\_Syn, PROT\_GO, PROT\_Conserve, PROT\_SSC, PROT\_Fold and DDI\_Interact. Model performance is evaluated through similarity metrics such as MACCS, RDK, MORGAN, Meteor and Levenshtein distance.
   % \item Byte-level prediction tasks contain . The performance is evaluated through METEOR, ROUGE-L and BLEU-1.  
\end{itemize}

% Additionally, we include composite scores for overall performance and metrics for hallucination detection and refusal rates in safety-critical queries.

\textbf{5. Tool calling API. }
Candidate tools were sourced exclusively from ToolUniverse\cite{gao2025tooluniverse}, which standardizes tool specifications in JSON format (including name, description, parameters, and execution endpoints). The selected tools for this evaluation were:

\begin{itemize}
    \item $ADMETAI\_predict\_solubility\_lipophilicity\_hydration$: A graph neural network-based platform for rapid prediction of ADMET properties across large chemical libraries.
    The output values of $Solubility\_AqSolDB$ and $HydrationFreeEnergy\_FreeSolv$ property represent results of MOL\_Solubility and MOL\_Freesolv task, respectively.
    \item $ADMETAI\_predict\_BBB\_penetrance$: used to execute MOL\_BBBP task, representing with $BBB\_Martins$ property value.
    \item $ADMETAI\_predict\_stress\_response$: used to execute MOL\_TOX task, representing with $SR-ARE$ toxicity value.
\end{itemize}

% This BioMol-LLM-Bench addresses key limitations in prior benchmarks, such as lack of multi-scale diversity and tool integration, providing a robust platform to gauge LLM progress in bio-chemistry.
% It facilitates direct comparisons across models and highlights areas for improvement such as protein understanding under amino acid level. 
% The dataset and evaluation scripts will be publicly released to support ongoing research in AI-driven molecular discovery.

% \textbf{Manuscripts may reference figures (e.g. Figure 1), tables (e.g. Table 1). Please do not include internal hyperlinks to elements within the manuscript as these are typically removed and replaced with our own linking when typesetting.}

\subsection{LLM Evalution}

\textbf{1. Model Selection and Deployment.}
Both open-source and closed-source LLMs were selected to enable transparent and reproducible deployment.
The primary models evaluated were listed in Table \ref{tab:tab3}.
%, including both general-purpose LLMs and fine-tuned domain-specific models. 
The parameter scale, model architecture, and thinking mode of these models vary.
For open-source models, we deployed them with $vllm$\cite{kwon2023vllm}.
% For large models such as Nature-LM, eight NPUs were needed.
For closed-source models, batch testing was conducted by calling the API interface ($openai.client$) with equivalent prompts.
Temperature was set to 0.7, and top-p was fixed at 1.0 by default.
Various comparative experiments and analyses were conducted, including:
\begin{itemize}
    \item The performance of general-purpose LLMs and domain-specific fine-tuned models on BioMol-LLM-Bench.
    \item Based on the same backbone model, compare model performance before and after CoT fine-tuning (Qwen-3-14B vs. DeepSeek-R1-Distill-Qwen-14B).
    \item Compare the performance of models with different parameter scales.
    \item The performance of different architecture models (dense transformer, MoE, Mamba) under similar parameter scales.
    \item The impact of input prompt complexity, think mode, and other configurations.
    \item The impact of external tool assistance on model performance.
\end{itemize}

\textbf{2. Prompt definition.}
Consistent prompts were designed to standardize input for general-purpose LLMs.
Prompts were stored as reusable templates in a configuration file.
For each task, a natural language prompt template was defined, incorporating the molecular representation and task-specific instructions.
Models were prompted in a standardized zero-shot manner.
%, example of detailed prompt template for $MOL\_Freesolv$ task:
To compare the prompt complexity influence on model performance, we also define the simplified prompt template for each task. 

% \needspace{10cm}
\begin{mdframed}[frametitle={Simple prompt template for MOL\_Freesolv task:}, backgroundcolor=white, leftmargin=5pt, rightmargin=5pt, skipabove=5pt, linewidth=1pt]
Context: Given the [SMILES] sequence of a molecule, your task is to predict its free energy. 

[SMILES]: \{\} 

Your response should be in the following format: 

Answer: \{your answer, representing with a floating-point number.\}
\end{mdframed}

\begin{mdframed}[frametitle={Detailed prompt template for MOL\_Freesolv task:}, backgroundcolor=white, leftmargin=5pt, rightmargin=5pt, linewidth=1pt]
Instruction: The solvation free energy of a molecule is a thermodynamic measure of how favorably a molecule dissolves in a solvent (typically water). 

Context: Given the [SMILES] sequence of a molecule, your task is to predict its free energy. 

[SMILES]: \{\} 

Your response should be in the following format: 

Explanation: \{your explanation for your answer, this is optional.\}

Answer: \{your answer, representing with a floating-point number.\}

Confidence: \{your confidence score between 0\% and 100\% for your answer.\}
\end{mdframed}

For fine-tuned domain-specific models such as NatureLM-8x7B and TxGemma-Chat-9B, since they have predefined fine-tuning templates, we construct prompts for each task by referencing the original template. %For NatureLM-8x7B, the prompt is,

% \needspace{10cm}
\begin{mdframed}[frametitle={Prompt template for NatureLM-8x7B:}, backgroundcolor=white, leftmargin=5pt, rightmargin=5pt, linewidth=1pt]
Instruction: Predict solvation free energy of the molecule. \{\} 

Response:
\end{mdframed}

% For TxGemma-Chat-9B, the prompt is,
% \needspace{10cm}
\begin{mdframed}[frametitle={Prompt template for TxGemma-Chat-9B:}, backgroundcolor=white, leftmargin=5pt, rightmargin=5pt, linewidth=1pt]
Instructions: Answer the following question about molecular property.

Context: The solvation free energy of a molecule is a thermodynamic measure of how favorably a molecule dissolves in a solvent (typically water). 

Question: Given the [SMILES] sequence of a molecule, predict its free energy. 

[SMILES]: \{\} 

Answer:
\end{mdframed}

\section*{Availability and implementation}
\label{sec:availability}
Source code is available at \url{https://github.com/AI-HPC-Research-Team/BioMol-LLM-Bench}, and the benchmark dataset is available at \url{https://github.com/AI-HPC-Research-Team/BioMol-LLM-Bench/tree/main/dataset}. The code is released under Apache-2.0 License.

% Source code is available at \url{https://github.com/AI-HPC-Research-Team/BioMol-LLM-Bench}, and the evaluation dataset is available at \url{https://github.com/AI-HPC-Research-Team/BioMol-LLM-Bench/tree/main/dataset}. The code, including model deployment, ToolUniverse integration, prompt templates, automated parsing, metric calculation and ablation studies, is released under the MIT License. The evaluation dataset, including raw model outputs, parsed tool calls, execution logs, ground-truth comparisons and performance metrics, is released under the Apache-2.0 License.

% \section*{Acknowledgments}
% This work was supported by the National Science Foundation under Grant No. XXXXXXX. We acknowledge the Computational Resources Center at University of Technology for providing GPU infrastructure. We thank the anonymous reviewers for their valuable comments and suggestions.

% \section*{Author contributions}

% \section*{Funding}

% \section*{Competing interests}
% The authors declare no competing interests.

\bibliography{sn-bibliography}% common bib file
%% if required, the content of .bbl file can be included here once bbl is generated
%%\input sn-article.bbl

\begin{appendices}

\section{Supplementary Table.}
\label{app:descript}

\begin{longtable}{p{1.2cm}p{3.2cm}p{10.5cm}}
    \caption*{Supplementary Table: Detailed descriptions of benchmark task.}\label{tab:appt1}\\
    
    \toprule
    \textbf{Level} & \textbf{Task (abbr.)} & \textbf{Description} \\
    \midrule
    \endfirsthead
    
    % \multicolumn{3}{c}%
    \multicolumn{3}{r}%
    % {\tablename\ \thetable\ -- \textit{Continued from previous page}} \\
    {\textit{Continued from previous page}} \\
    \toprule
    \textbf{Level} & \textbf{Task (abbr.)} & \textbf{Description} \\
    \midrule
    \endhead
    
    \midrule
    \multicolumn{3}{r}{\textit{Continued on next page}} \\
    \endfoot
    
    \bottomrule
    \endlastfoot
    
    % \begin{tabularx}{\textwidth}
        % \toprule
        % Level & Task (abbr.) & Description \\
        % \midrule
        L0 & General\_Text  & Domain knowledge understanding with text-input only. \\
        \midrule
        L1 & MOL\_Solubility & The solubility of a molecule refers to its ability to dissolve in a solvent (typically water) to form a homogeneous solution. \\
         & MOL\_Freesolv & The solvation free energy of a molecule is a thermodynamic measure of how favorably a molecule dissolves in a solvent (typically water). \\
         & MOL\_Syn & Predict the forward synthesis production of two molecules. \\
         & MOL\_Resyn  & Given production, predict the reactant molecules.  \\
         & MOL\_Excited & E1 property of a moleculer refers to the energy of lowest singlet excited state relative to the ground state. Predict the energy of lowest singlet excited state relative to ground state of molecule in eV units.  \\
         & MOL\_Thermo & Predict the heat capacity of molecule in kcal/mol/K scale.  \\
         & MOL\_BBBP & The BBBP (Blood-Brain Barrier Penetration) property represents whether a molecule can cross the blood-brain barrier.  \\
         & MOL\_HIV & The HIV replication prevention property refers to a molecule's ability to inhibit one or more steps in the HIV replication cycle, thereby preventing viral proliferation. \\
         & MOL\_HLGap &  The HOMO-LUMO gap (Highest Occupied Molecular Orbital - Lowest Unoccupied Molecular Orbital gap) represents energy difference between HOMO and LUMO (measured in eV or kcal/mol). \\
         & MOL\_TOX &  The SR-ARE (Sulfonamide Reactive - Aryl Ester) toxicity refers to the adverse effects caused by molecules containing sulfonamide and aryl ester functional groups. \\
        \midrule
        L2 &PROT\_Melt & The melting point of a protein is the temperature at which 50\% of the protein unfolds, marking the midpoint of the transition from the folded (native) state to the unfolded (denatured) state.  \\
         & PROT\_Energy & The stability of a mutated protein sequence represents the free energy difference between mutated sequence and its native sequence.  \\
         & PROT\_Fitness &  The fitness landscape represents how mutations affect functionality (e.g., enzymatic activity, binding affinity, stability) of a protein.  \\
         &PROT\_Mutation  & The mutation score of mutated protein quantifies the functional/structural impact of a small fraction of amino acid changes. \\
         &PROT\_Conserve  & The conservation score measures how evolutionarily conserved a residue across homologs (reflects purifying selection). \\
         & PROT\_Location &  The subcellular localization class of a protein refers to its specific compartment or organelle within a cell where it predominantly resides and functions. \\
         & PROT\_EC  &  The Enzyme Commission (EC) system classifies enzymes into a hierarchical numbering system (e.g., EC 1.1.1.1 for alcohol dehydrogenase) based on their catalytic reactions. The task is to predict first-level EC class number of protein.\\
         & PROT\_GO  &  The Gene Ontology (GO) describes protein functions in three categories: Molecular Function (MF), Biological Process (BP) and Cellular Component (CC). The task is to predict the Molecular Function (MF) properties of protein.  \\
         & PROT\_SSC  & Secondary structure classification typically follows the DSSP (Define Secondary Structure of Proteins) standard, which categorizes each amino acid of the protein into distinct classes.  \\
         & PROT\_Fold &  Predict the 3D structure of protein. \\ 
        \midrule
        L3  & PPI\_Type  & Predict the PPI interaction types between two proteins. \\
        & PPI\_Binary  & The task of predicting whether two proteins interact or not. \\
        & DDI\_Interact & Predict the DDI interaction type between two drugs. \\
        & PLI\_BA  & Predict the PLI binding affinity between protein and ligand. \\
        & CI\_AbAg &  Predict the AbAg binding affinity between antibody (heavy and light chain) and antigen.  \\
        % \bottomrule
    % \end{tabularx}
\end{longtable}

\end{appendices}

\end{document}